\documentclass[12pt]{scrartcl}
\usepackage{gensymb}
\usepackage{caption}
\usepackage{subcaption}
\usepackage{svg}
\usepackage{scalerel}
\usepackage{gensymb}
\usepackage{makecell} 
\usepackage{pifont}
\newcommand{\checkmark}{\ding{51}}
\newcommand{\xmark}{\ding{55}}
\usepackage{rotating}

\makeatletter
\def\@xfootnote[#1]{%
  \protected@xdef\@thefnmark{#1}%
  \@footnotemark\@footnotetext}
\makeatother

\usepackage[margin=1in]{geometry}
\usepackage{graphicx}
\usepackage{caption}
\usepackage{subcaption}

\usepackage[utf8]{inputenc}

\usepackage[english]{babel}

\usepackage{setspace}
\usepackage{indentfirst}

\usepackage{appendix}

\usepackage{amssymb}
\usepackage{amsmath}
\usepackage{amsthm}
\usepackage{mathrsfs}
\usepackage{bm,bold-extra}

\usepackage{hyperref,xcolor,bookmark}
\hypersetup{pdfborder = 0 0 0}
\bookmarksetup{numbered,open}
\definecolor{ourred}{HTML}{C5050C} 
\definecolor{ourblue}{HTML}{3F547D} 

\usepackage{changepage,pdflscape,afterpage,graphicx,boxedminipage}

\usepackage{authblk}
\usepackage{caption,array,verbatim,scrextend,enumitem}

\usepackage{authblk}

\usepackage{longtable,threeparttablex,multirow,multicol,booktabs,dcolumn}

\usepackage[style = authoryear,
            citestyle = authoryear,
            backend=biber,
            doi=false,
            isbn=false,
            eprint=false,
            url=false]{biblatex}

\usepackage{multirow}

\title{Alignment Helps Make the Most of Multimodal Data}

\date{\today}

\begin{document}

\doublespacing

\begin{center}
    \begin{LARGE}
    \textbf{Alignment Helps Make the Most of Multimodal Data\footnote[*]{Thomas Gschwend, Achim Hildebrandt, Frederik Hjorth, Stefan Müller, Oliver Rittmann, Joseph Ornstein, Wang Leung Ting, and Michelle Torres provided tremendously helpful comments to earlier versions. We thank Daniel Kuhlen for his excellent research assistance. The authors acknowledge support by the state of Baden-Württemberg, Germany, through bwHPC. Earlier versions of the draft were presented at the 5th Annual COMPTEXT Conference, Glasgow, 12-13 May 2023, at the 13th Annual EPSA Conference Glasgow, 22-24 June 2023, the ECPR SGoP Conference, Vienna, 6-8 July 2023, the Political Studies Association of Ireland (PSAI) Annual Conference, 20-22 October 2023, Belfast, the 46th Annual Conference of the IASGP, Mannheim, 30 November-1 December 2023, the 81st MPSA, Chicago, 4-7 April 2024, and the Text-as-Data Reading Group, 8 May 2024, Online.}\\}
    \end{LARGE}
    \vspace{0.4cm}
\begin{large}
    Christian Arnold\textsuperscript{1} and Andreas Küpfer\textsuperscript{2}\break
  \textsuperscript{1}University of Birmingham, \url{c.arnold.2@bham.ac.uk}\\
  \textsuperscript{2}Technical University of Darmstadt, \url{andreas.kuepfer@tu-darmstadt.de}\break\break
 April 11, 2025
 \end{large}
\end{center}

\begin{abstract}
Political scientists increasingly analyze multimodal data. However, the effective analysis of such data requires aligning information across different modalities. In our paper, we demonstrate the significance of such alignment. Informed by a systematic review of 2,703 papers, we find that political scientists typically do not align their multimodal data. Introducing a decision tree that guides alignment choices, our framework highlights alignment’s untapped potential and provides concrete advice in research design and modeling decisions. We illustrate alignment’s analytical value through two applications: predicting tonality in U.S. presidential campaign ads and cross-modal querying of German parliamentary speeches to examine responses to the far-right AfD.

\end{abstract}

\newpage

\section{Introduction}\label{introduction}

Over the past two decades, political science has made tremendous advances in learning how to work with unstructured data, be it text \autocite{laver2003extracting}, audio \autocite{knox_lucas_2021}, or images \autocite{Torres2024}. While analyzing these modalities separately allowed for important insights, their combination is particularly promising: Many human activity relies on more than one modality alone. Anyone posting a text and an accompanying photo on social media wants to convey a message that requires both. Face-to-face communication is more than just the words of a transcript alone; it can include non-verbal signals. A researcher focusing on, for example, just the text of a social media post or just the transcript of a political speech could miss out on important information.

But modeling such multimodal data comes with a key challenge: How does the information from one modality relate to relevant information in the other? This alignment is not a trivial problem since the data may look very differently\textemdash compare how distinct the data representation of a hand clap is in image data (pixels) and audio data (waves). In addition, information may relate to each other at conceptually different levels. A hand clap is directly correlated in the audio and visual modalities. Yet, a sarcastic statement's wink of an eye requires a much deeper semantic understanding of the data.\footnote{We are using the term \emph{semantics} not simply to represent meaning in language, but also in other modalities.} Finally, signals from different modalities may interact with one another in various ways. A specific tonality in a speaker's voice might reinforce a statement or, in the case of sarcasm, would invert it.

A key requirement for overcoming these obstacles is to carefully align the modalities. It is not enough to just offer models the information from both modalities, for example, by syncing a video's audio and visual data based on time alone or concatenating embeddings from two modalities. To make the most of multimodal information, researchers should carefully consider what the respective elements of interest are in each modality (e.g., words, sounds, objects) and define how they relate to one another. Only once they know how they intend to align the data should they analyze it. 

Choosing the apt technical solution for alignment depends on three core questions. Is the data continuous, or does it already contain discrete semantic elements? In the case of the latter, does the data represent the elements in a semantic or non-semantic way? Finally, is alignment a separate model and occurs as pre-processing of the data before the modeling \autocite{denny2018text}? Or is alignment implicitly built into the model itself?

We show the potential of aligned data in two applications. In the first case, we predict the tonality of TV ads from the presidential election campaign of the USA in 2020. Aligning the information from audio, image, and text allows the model to make the most of the information in the data and improves model performance. In the second case, we align audio-visual data from MPs' speeches in the German \textit{Bundestag}. With connections between the semantic elements of the modalities, it is then possible to query one modality for information in another. Using such a cross-modal query, we study how German MPs react to the arrival of the far-right AfD.

The contribution of our paper is straightforward. We define alignment, explain its importance for multimodal data analysis, and outline the technical steps for its implementation. In the first section, we introduce the core challenges of multimodal data\textemdash heterogeneity, connectedness, and interaction\textemdash and show how alignment helps address them. The second section reports results from a systematic literature review of 2,703 political science papers. We find only two cases where scholars have aligned their multimodal data for analysis. In the third section, we offer practical guidance and introduce a decision tree that helps implement alignment in practice. Spotlighting its potential for research, the fourth section showcases how to use alignment for performance improvement and cross-modal queries. The final section concludes.

\section{Addressing the Challenges of Multimodal Data with Alignment}

A modality is a way to describe a phenomenon through a sensor \autocite{baltruvsaitis2018multimodal, barua2023systematic}. Multimodal data represents the same phenomenon through different sensors. Not all of them record the same aspect of the phenomenon, and the union of their information offers a more comprehensive description than one modality alone \autocite{liang2023foundations}. For example, a social media post may have two modalities: text and an image. Two cameras that take a photo of the same event from different perspectives are also multimodal data. Humans naturally learn to perceive the world through various senses when they grow up. But, while models can tap into information that resembles the human experience of the world, they require direction on how to integrate the information from the respective modalities. This is where alignment plays a crucial role.

\subsection{Three Challenges of Multimodal Data Analysis}
Three core challenges must be overcome to model multimodal data \autocite[5ff.]{liang2023foundations}. To begin, multimodal data is heterogeneous. Modalities typically differ in how they represent information. Language data is built from letters and words; audio data is a time series of oscillations; a color image is a three-dimensional object of pixel values across height, width, and color band; and video data adds time as another dimension. The individual data representations also relate these information bits to one another in various ways. The grammar of a sentence works hierarchically; an image represents information spatially, and audio and images in a video are sequences. Moreover, modalities may be heterogeneous in their capacity to represent a phenomenon. For example, recent studies have found that the audio signal might be quite relevant to identify emotions in political speeches beyond `just' text alone \autocite{dietrich2019pitch, knox_lucas_2021, rittmannforthcominglegislators}. The modalities can also have different noise levels; think of typos or bad cameras.

Despite its heterogeneity, multimodal data is connected, as the various data streams describe the same event. This connection can occur at different semantic levels. A hand clap, for example, is easily identifiable in both the audio and visual modalities. In contrast, a politician’s sarcastic social media post\textemdash where a winking photo accompanies text\textemdash requires a deep understanding of the image and the text, as well as how their combination creates meaning \autocite{ashyley2017sarcasm, leicht2023nightly}.

\begin{figure}
    \centering
    \includegraphics[width=0.6\linewidth]{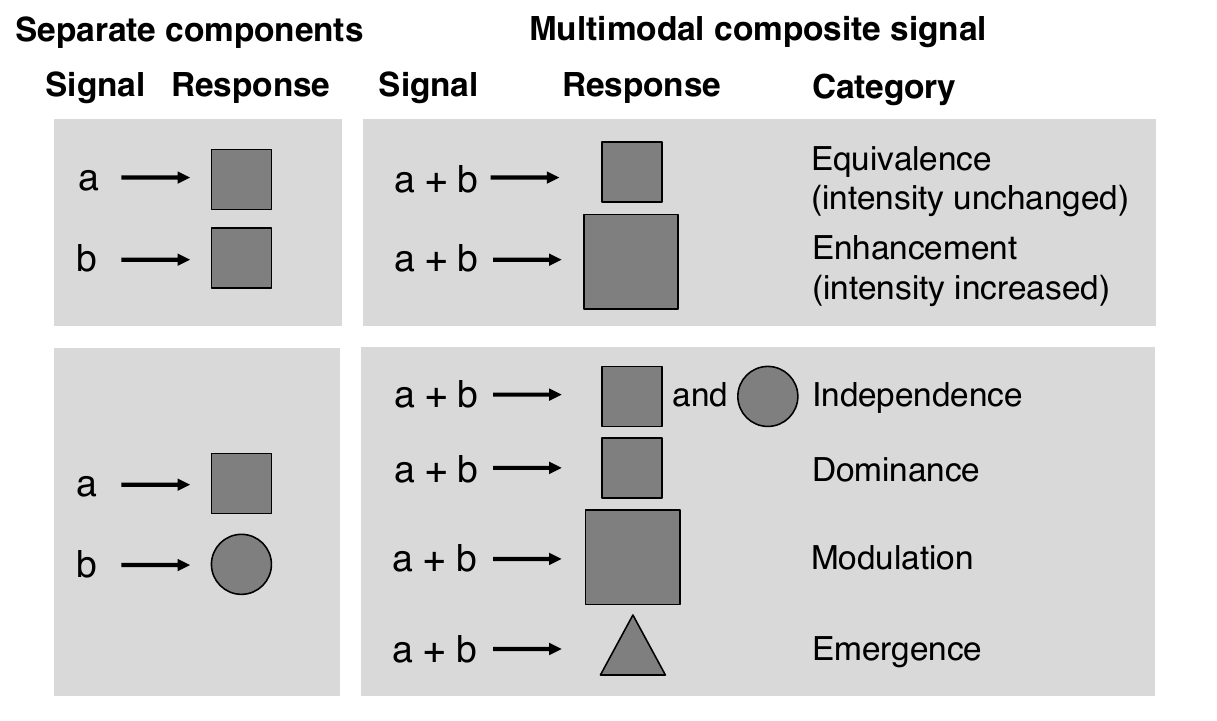}
    \caption{Different outcomes when combining redundant and non-redundant multimodal signals (based on Partan and Marler \autocite*{partan99}).}
    \label{fig:combiningMMD}
\end{figure}

Lastly, combining the information from the modalities can occur in all possible ways (Figure~\ref{fig:combiningMMD}). Even if the signals in the respective modalities are mostly redundant, their combination might lead to the same response or enhance their intensities. The tonality of the adjective in ``This policy proposal is \emph{outrageous}'' can make a key difference in deciding whether the sentence is more factual or emotional. If the signals are nonredundant, there are even four possible scenarios. The two signals may be independent. The facial expression of a person in the background of an MP may be unrelated to their speech. One signal may dominate the other, e.g., when the content of the speech supersedes erratic hand movements. Signals can modulate each other. Just imagine how crucial the wink of an eye is to understanding sarcasm. Finally, a new meaning might emerge. In a social media post, dog-whistle communication would use an innocuous image, like \textit{Pepe the Frog}, paired with ambiguous text, such as ``Feeling froggy today,'' to convey hidden ideological messages.

\subsection{What Is Alignment?}
To understand alignment, we look more closely at how elements in each modality jointly create meaning. The data in each modality consists of a set of elements that humans can interpret. For example, the elements in a social media text are the respective words. An audio file can be transcribed to generate words, but it can also contain a particular sound of interest, like a clap. In an image, the elements can be objects, such as a specific person. In the visual modality from a video, this object would move through time. Alignment identifies the connections between elements across multiple modalities \parencite{baltruvsaitis2018multimodal, liang2023foundations}. These connections do not have to be one-to-one but include all possible mappings like one-to-many (e.g. the word ‘victory’ and an image with a raised fist, a waving flag, and a smiling politician), many-to-one (e.g., the words ‘heatwave,’ ‘record temperatures,’ and ‘scorching sun’ and an image of a cracked, dry landscape) and many-to-many (e.g., the words ‘protest,’ ‘demands,’ and ‘police presence’ and an image of a crowd of demonstrators, protest signs, and a police barrier.)

Some examples further the understanding of this definition of alignment. In an emotional political speech, the semantics of the words and the non-verbal communication both matter. A model tasked with determining the emotions benefits from aligned information across the modalities because it connects facial expressions and body language in the visual modality, particular voice colors in the audio modality, and the semantic meaning of particular words. 

Connecting the audio and visual modality of a video on time alone is not alignment in the sense of our definition. Unless researchers define what they expect to find in those modalities, such as particular sounds, words, or gestures, no distinct elements are connected. In a second negative scenario, a researcher seeks to combine a text and an image modality for social media analysis based on the respective pre-trained embedding representations. While it is technically possible to concatenate the embeddings and analyze this larger embedding with a neural net, the modalities would still not be aligned. There is no connection between the semantic representation of the elements in each embedding space, something a projection into a joint latent space could provide.

\subsection{How Alignment Helps Address the Challenges of Multimodal Data}

Alignment is crucial to address the three challenges in multimodal data analysis.

It tackles the heterogeneity of multimodal data by relating relevant information to one another, even if the data looks very different. Without alignment, it remains unclear what parts of the data in one modality relate to what parts of the data in another. Alignment also specifies how to connect the information from the respective modalities. This might be straightforward and evident if signals from multiple modalities are linked, such as a hand clap's visual and audio signals. However, alignment is more challenging when the connection requires deep semantic understanding, e.g., for detecting sarcasm or dog whistles. Ultimately, alignment also addresses the combination problem, as analysts must carefully consider how the signals from the modalities may interact. This has consequences not only for the modeling but can already affect data collection. When annotating data for a research project on parliamentary speeches, it might be enough to study the transcript alone. However, if nonverbal communication plays an important role, then including the other modalities is critical to avoid information loss or omitted-variable bias.

Alignment has many positive consequences. Integrating information from different modalities enhances model performance, whether for descriptive, predictive, or causal inference purposes \autocite{liang2023multiviz}. Aligned modalities may compensate for noisy or missing data in individual modalities, which is why it is particularly promising in low-resource modalities \autocite{Liang21, liang2023foundations}. In addition, the established connections allow for cross-modal queries. An analyst can search for information in one modality and retrieve relevant information from another \autocite{baltruvsaitis2018multimodal, Liang21}. For example, politicians may rely on particular vocal pitches or facial expressions when using specific keywords such as \emph{immigrant} or \emph{climate change}, or they could use particular phrases to react to acoustic (booing, clapping) or visual cues (facial expressions, hand gestures).

\section{How Do Political Scientists Align Modalities?}

\begin{figure}[h]
    \centering
    \includegraphics[width=.8\textwidth]{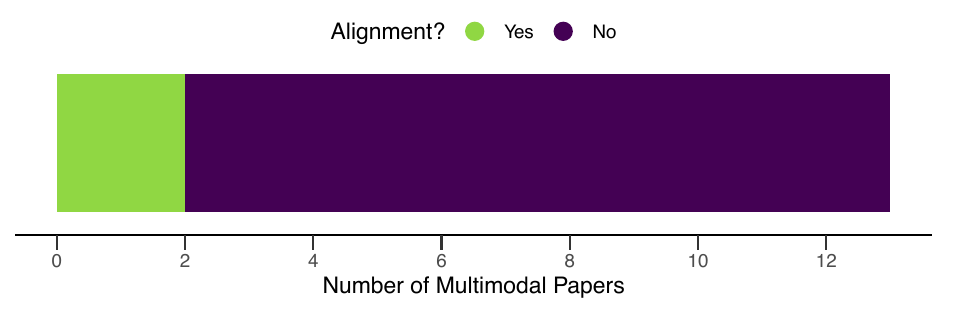}
    \caption{Do political scientists currently align? Hand annotation of all publications analyzing multimodal data in AJPS, APSR, BJPS, JOP, PA, and PolCom between 2019 and 2024.}
    \label{fig:lit_review_alignments}
\end{figure}

We implemented a systematic literature review to understand how political scientists have recently been aligning multimodal data. Scraping Google Scholar, we identified all articles in the American Journal of Political Science, American Political Science Review, British Journal of Political Science, The Journal of Politics, Political Analysis, and Political Communication that a) featured the keywords \texttt{multimodal OR video OR audio OR text OR image} and b) have been published over the last five years: overall 2,703 papers. Manually annotating all articles, we found that only 13 use multimodal analysis: six in PA, three in APSR, three in PolCom, and one in BJPS. Lastly, we annotate whether the articles align modalities for their analyses. According to the definition, alignment requires a) defining the elements in each modality, b) their relationship, and c) explicitly or implicitly accounting for the relationship when modeling the data. Figure~\ref{fig:lit_review_alignments} shows the result of our systematic literature review. Out of the 13 multimodal papers we identified, only two implement alignment.\footnote{Please see the Appendix for the full list of papers, annotation details, and further analyses.}

\section{Modeling Alignment as Part of the Data Generating Process}

When modeling tabular data, social scientists are often guided by theoretically informed ideas about the data generating process \autocite{Aldrich08, brauninger2020theory}. These principles have recently been extended to unstructured data formats like text \autocite{egami2022make, slapin2008scaling}, or audio \autocite{knox_lucas_2021}. While machine learning approaches are often less concerned in this regard \autocite{grimmer2021machine}, thoughts must go into understanding the data and its structure, and alignment should be part of these considerations. 

\begin{figure}[h]
    \centering
    \includegraphics[width=0.75\linewidth]{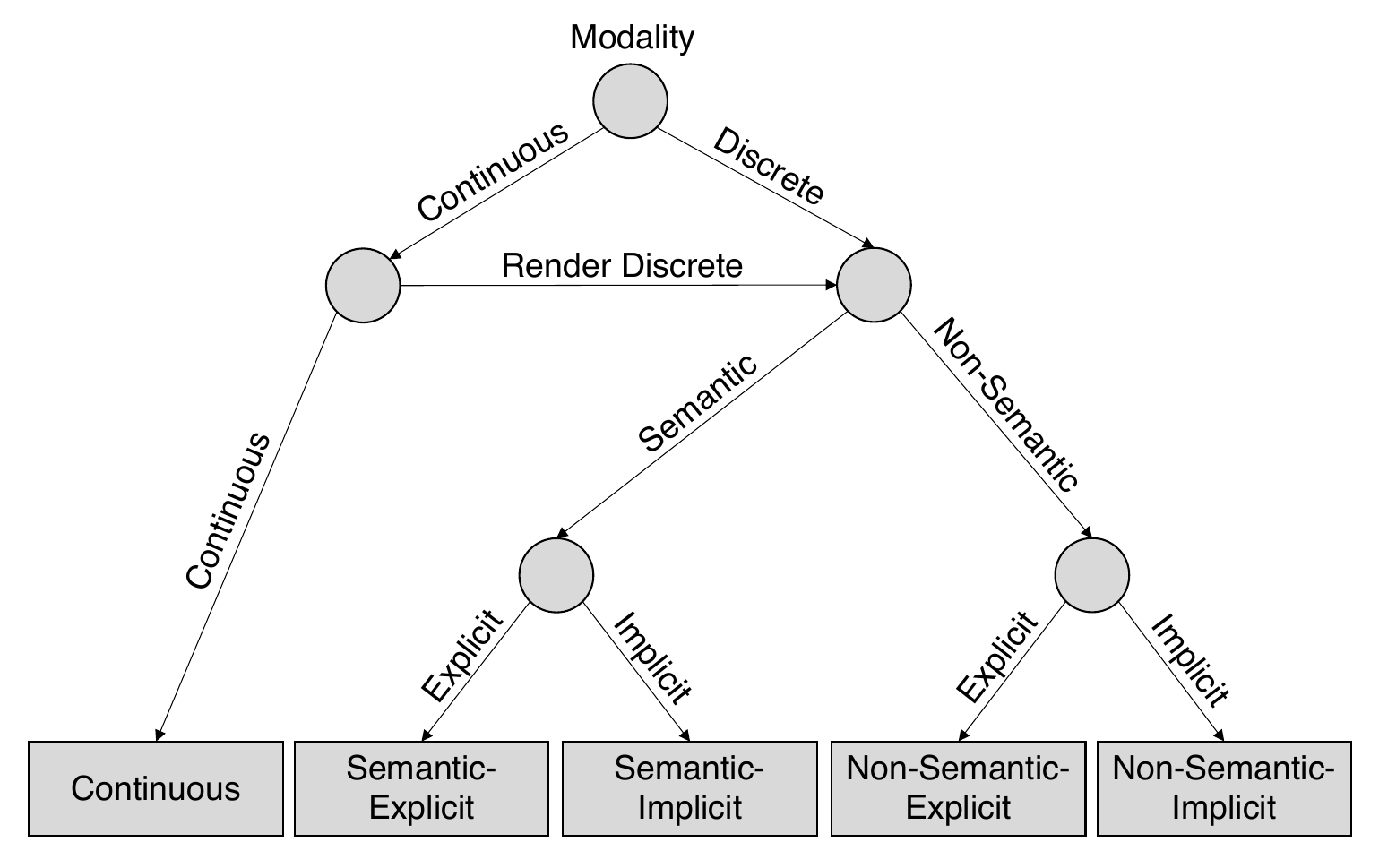}
    \caption{Decision tree for modeling alignment. Decisions are based on the data structure and the overall model architecture.}
    \label{fig:decisiontree}
\end{figure}

A first important distinction is to understand whether the data in a modality is continuous\textemdash think pixels in an image or audio waves\textemdash, or whether the modality already consists of discrete elements, such as words in a text modality or categories in tabular data (Figure~\ref{fig:decisiontree}). 

If the data is continuous, should the semantic elements be extracted? It is possible to transcribe the words from audio data or identify any other sound of interest. Rask \autocite*{rask23}, for example, uses aligned text-audio data to analyze parliamentary speech. Likewise, algorithms have been used in political science to identify the pixels that belong to particular objects in an inductive \autocite{Torres2024} and deductive way \autocite{scholz2024improving}. Continuous data could also remain in its rather unstructured shape without explicitly identifying elements. 

In the case of data with discrete elements, the next thought should be about how the data is represented. Embeddings have proven to be a powerful tool for encoding the semantics of discrete elements in unstructured data. In political science, they have been used to demonstrate how the meaning of concepts shifts over time and space \autocite{Rodman_2020, RODRIGUEZ_SPIRLING_STEWART_2023}. A more traditional way of representing a text or image corpus remains non-semantic. Term-document-matrices \autocite{laver2003extracting, slapin2008scaling} or object-image-matrices \autocite{scholz2024improving, Torres2024} record whether or how many instances of each element occur in each of the corpus' documents.

If the data is continuous, researchers have mainly two different options for modeling alignment \autocite{liang2023foundations}. In adversarial training, the goal is to find data representations that are so similar that a classifier cannot identify which modality data is coming from. This has, for example, been applied to identify similar actions in different video data \autocite{munro2020multi}. Another approach, dynamic time warping, aligns sequential data by adjusting temporal structures to maximize similarity, with extensions incorporating canonical correlation analysis to jointly optimize alignment \autocite{trigeorgis2017deep}. 

If the data represents discrete elements, researchers can choose between two modeling paradigms, explicit and implicit alignment. 
In the former case, alignment occurs with a distinct model focusing on aligning the elements across modalities, e.g., with a supervised classifier. Alignment is then a form of pre-processing the data \autocite{denny2018text}. Any further analyses build on this aligned data and require a separate model. 
In implicit alignment, the model has a substantively different purpose, such as detecting sarcasm or humour \autocite{pramanick2022multimodal}. However, its architecture enables it to align different modalities and improve performance when trained to achieve its intended goal.

Aligning semantic representations is similar to what political scientists do when they map latent spaces onto one another \autocite{aldrich1977method, Enamorado_López-Moctezuma_Ratkovic_2021}. The key difference is that while political scientists have traditionally focused on stretching and scaling in one-dimensional settings, aligning embeddings requires coordinating a high-dimensional space. Discovering semantic representation has led to a tremendous performance increase when working with unstructured data \autocite{socher2013zero, weston2011wsabie}. Unsurprisingly, key advances in multimodal analysis are founded on coordinating the respective semantic spaces \autocite{barua2023systematic, liang2023foundations}. Recent generative models can generate images and videos based on text or generate text based on audio prompts \autocite{girdhar2023imagebind}.

In explicit approaches, models accomplish alignment with unsupervised methods, such as adversarial learning in auto-encoders \autocite{trigeorgis2017deep}, or by integrating canonical correlation analysis with neural network architectures \autocite{andrew2013deep}. Optimal transport-based approaches are another solution since they seek to achieve alignment through divergence minimization. Lupu, Selios, and Warner \autocite*{lupu2017new} have used such approaches to match the political attitudes of individuals and political elites. 

Aligning embeddings implicitly requires incorporating the alignment step into the model that ultimately implements the analysis. Transformer architectures that rely on cross-modal self-attention have proven helpful in this context. The transformer architectures dynamically direct a model's attention across modalities, even if the relevant pieces of information occur at different time steps \autocite{lu2019vilbert, tsai2019multimodal}. Scholars have used such architectures in political science to study political texts across languages \autocite{Licht_2023}.

However, multimodal analysis does not necessarily have to build on semantic representation. 
Explicit alignment of non-semantic representations enables more direct connections between modalities. Traditionally, supervised models establish these links by labeling elements within each modality, including cases that account for complex relationships such as one-to-one, one-to-many, or many-to-many mappings \autocite{gebru2017audio, hu2016natural}. A typical example involves associating objects in an image with descriptive words that capture the scene’s context \autocite{cirik2020refer360, hu2016natural}. Assessing the quality of such explicit alignment can already be relevant information, e.g., to detect deepfakes in TV news \autocite{agarwal2023watch}. 

While modern models using implicit alignment usually rely on semantic representations, older models implement non-semantic implicit alignment. Late fusion architectures that merge information about elements at a late stage in the model must coordinate the information about the elements in each modality in the early stages \autocite{atrey2010multimodal}. Hidden Markov Models, for example, have been used to segment and align audio and text modalities \autocite{sjolander2003hmm}.

\section{The Potential of Alignment in Practice}

We now showcase the potential of alignment in practice. In one application, we improve the prediction of the tone of campaign videos from the 2020 U.S. presidential election, i.e., whether they are positive or negative ads. Here, we spotlight the implicit alignment of semantic representations to efficiently use multimodal information. In a second application, we study the German \textit{Bundestag} and how parties react to the arrival of the far-right party \textit{Alternative für Deutschland} (AfD). We identify discrete elements (pitch per word in audio modality, gazing at the AfD in visual modality) and explicitly connect modalities based on when these elements occur. Here, the alignment allows us to understand whether MPs are more upset when they address the AfD.

\subsection{Alignment to Improve the Prediction of Tone in Campaign Videos}

For parties and politicians, campaign videos are essential to reach out to potential voters and differentiate themselves from their opponents. In negative campaigning, for example, political actors highlight the opponents' weaknesses, failures, or controversial actions rather than promoting their own policies or achievements \autocite{laupomper2001negative, Meirick2018truth, PATTIE2011333}. Voters perceive more negative ads as less accurate but exaggerate how centrist a candidate is \autocite{Meirick2018truth}. While humans can quickly capture a video's tone, it is not a trivial task for a machine. Automatically annotating campaign videos can serve as an important research tool for analyzing political ads at scale.

The Wesleyan Media Project provides access to 1,150 video advertisements from the 2020 U.S. Presidential campaign, including manually annotated variables \autocite{fowler_2023_2020}. Relying on the English language videos that last at least two minutes and have a positive or negative tone (1,077 videos), we annotate the tone of the videos with a deep neural network. We make the case for a multimodal approach. Information for automatically identifying tonality comes not from text alone, but often sits in visual and audio cues. To efficiently use the information from the respective modalities, our model architecture implicitly aligns the modalities while annotating the tonality of the videos.

The raw data from a video is a continuous representation of pixels and sound waves. Following Figure~\ref{fig:decisiontree}, we prepare the data in each modality for analysis. As a first step, we render the continuous modalities into discrete audio, visual, and text data. We automatically transcribe the text modality from the audio modality with \textit{WhisperX} \autocite{bain2023whisperx}. To capture non-verbal communication, we take small segments of the audio data every 0.1 seconds with an overlap of 0.05 seconds. We downsample the visual modality and take only every 8th frame. 

The next step is to represent each discrete element with an embedding vector. We use pre-trained \textit{GloVe} embeddings \autocite{pennington2014glove} for the text modality. For the audio snippets, we rely on \textit{pyAudioAnalysis} \autocite{giannakopoulos2015pyaudioanalysis} and extract 34 typical audio features. In the visual modality, we first identify whether each image contains a person. If so, we determine 170 facial features with \textit{Py-Feat}\footnote{https://py-feat.org/}.

\begin{figure}[h]
\centering
   \includegraphics[width=.7\linewidth]{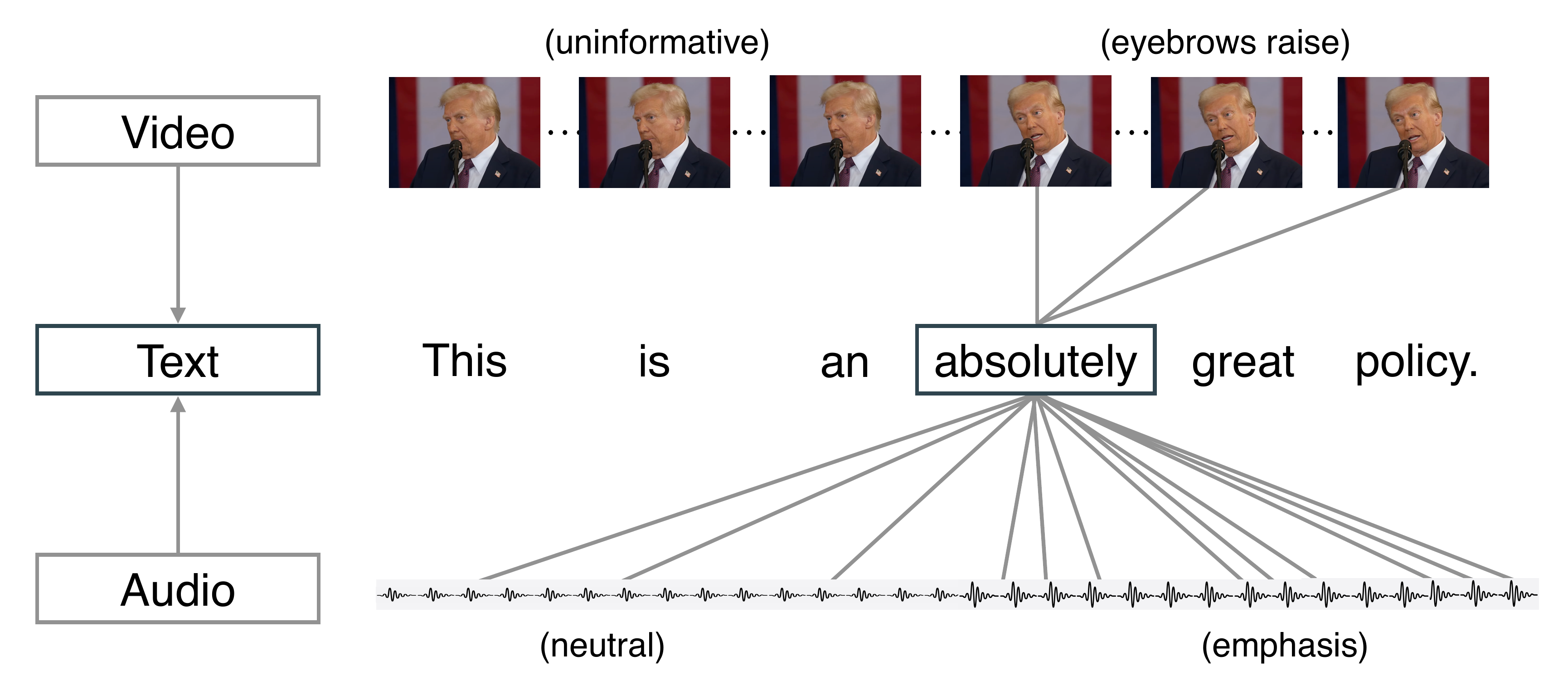}
 \caption{Implicit alignment using cross-modal transformers following Tsai et al. \autocite*{tsai2019multimodal}. Lines exemplarily show cross-modal attention weights between images and text as well as audio and text for the word 'absolutely'.}
 \label{fig:alignment_cross-modal}
 \end{figure}
 
Finally, we feed all three embedding representations into a neural net that uses cross-modal self-attention to align and extract relevant information when annotating tonality. Transformers weigh the relevance of input elements of a sequence to solve a model's task, i.e., the model learns to focus on more ``important'' parts when processing information \autocite{vaswani2017attention}. Self-attention on a single modality is a widely applied concept in unimodal settings (mainly text). Tsai et al. \autocite*{tsai2019multimodal} proposes an adapted transformer-based architecture that applies self-attention within and across all modalities.

Technically, this architecture implements several transformers for cross-modal attention. For each modality $X_T$ (Text), $X_V$ (Visual), and $X_A$ (Audio), we first pass the embeddings of all modalities through a temporal convolution layer, allowing the model to be aware of the neighboring modalities.\footnote{The model also stores positional embeddings to keep temporal information of all modalities.} Each modality in the dataset requires a dedicated transformer block for every other modality it connects with, e.g., $X_T$ requires $X_T \rightarrow X_V$ and $X_T \rightarrow X_A$. These blocks behave similarly to transformers translating text into another language. However, cross-modal attention does not directly translate $X_T$ into $X_V$ but latently adapts the representations across modalities \autocite[6559]{tsai2019multimodal}. Finally, an additional self-attention transformer processes the concatenated outputs of all blocks for prediction.

Figure~\ref{fig:alignment_cross-modal} illustrates the merits of this implicit alignment approach. The model subtly focuses on sequential relationships between modalities, connecting information from text, visual, and audio modalities. For example, the self-attention mechanism associates the word \textit{absolutely} with related sequences in both vision and audio and vice-versa. The non-verbal communication of the audio and image modality helps disambiguate the words' semantic meaning.

Building on Tsai et al. \autocite*{tsai2019multimodal}, we implement an ablation study to study the effect of alignment and offer the model five different versions of the data: each modality individually, a concatenated version of the data without the transformer blocks, and, finally, the model that includes the transformers to align modalities implicitly. In each scenario, we split our video data into a training, validation, and test set and determine the optimal combination of hyperparameters using a grid search of predefined hyperparameters \autocite{Arnold_Biedebach_Küpfer_Neunhoeffer_2024}.\footnote{Please see Appendix for hyperparameter values (Table~\ref{tab:hyperparameter_ranges}).}

\begin{table}[h]
\centering
\caption{Out-of-sample F1 scores of tone classification in unimodal and multimodal scenarios with and without cross-modal attention. Results averaged over five random training/validation/test set combinations. Complete list of results per seed in Appendix Table~\ref{tab:appendix_results}.}
\label{tab:implicit_results}
\begin{tabular}{p{2.8cm}p{3cm}ccc}
\toprule
\multicolumn{2}{c}{\textbf{Multimodal F1 (SD)}} & \multicolumn{3}{c}{\textbf{Unimodal F1 (SD)}}\\
\cmidrule(lr){1-2} \cmidrule(lr){3-5} 
\multicolumn{1}{c}{\makecell[c]{\textbf{Cross-modal} \\ \textbf{Attention}}} & \multicolumn{1}{c}{\makecell[c]{\textbf{No cross-modal} \\ \textbf{Attention}}} & \multicolumn{1}{c}{\textbf{Text}} & \multicolumn{1}{c}{\textbf{Audio}} & \multicolumn{1}{c}{\textbf{Visual}} \\
\midrule
\multicolumn{1}{c}{0.945 (0.012)} & \multicolumn{1}{c}{0.920 (0.019)} & 0.931 (0.006) & 0.851 (0.000) & 0.856 (0.002) \\
\bottomrule
\end{tabular}
\end{table}

Table~\ref{tab:implicit_results} reports F1 Scores on holdout test sets. Unimodal models incorporating the audio (0.851) or visual (0.856) modality seemingly do not contain all the information necessary to distinguish between negative and positive tones. To give an intuition, an accuracy score of 0.85 would mean that roughly one out of seven classifications is going wrong. A model that uses text only (0.931) outperforms the other two modalities and is better than the multimodal approach without cross-modal attention (0.920). Again, to compare, at an accuracy of 0.93, a model is wrong in only one out of 14 cases. However, the cross-modal attention architecture (0.945) improves model performance by yet another bit, making more efficient use of the available data. From a user perspective, a corresponding accuracy score translates into about one wrong prediction out of 20.

In sum, we use alignment in this application to tackle in particular the heterogeneity of multimodal data. The transformer architecture allows the model to efficiently connect relevant information when identifying the tone of the videos.

\subsection{Alignment for Cross-Modal Queries in Parliamentary Speeches}

Europe has witnessed a rise of far-right parties in recent years \autocite[e.g.,][]{radical_betz_1995, ivarsflaten2008pop}. In Germany, the AfD increased the political spectrum represented in the \textit{Bundestag} with its ideological position at the extreme right \autocite{atzenpodien2022afd}. Current studies of the reactions of German MPs to the AfD focus on transcripts of parliamentary debates \autocite{schwalbach2023talking}. Given the relevance of non-verbal communication in such contexts \autocite{dietrich2019pitch}, we suggest including the audio and visual modality to analyze politics on the floor.

Are German MPs more upset when talking to the AfD? We study how MPs address the AfD in their speeches using all 15,553 video recordings of plenary debates held in the 19th German \textit{Bundestag} between 2017 and 2021, the first legislative period with the AfD.\footnote{We scrape the plenary debates from the \textit{Bundestag} homepage (\url{https://www.bundestag.de/parlamentsfernsehen}).} To answer our research question, we build two measurements from three modalities. Vocal pitch reliably measures arousal in political speech \parencite{dietrich2019pitch, knox_lucas_2021, rittmannforthcominglegislators}. We, therefore, calculate an average pitch per word for each MP and expect that MPs will speak with a higher voice than usual when they address the AfD. But when do MPs turn to the AfD? 
An MP who speaks to the far right will look at them. Given the layout of the parliamentary floor in the Bundestag, measuring the MPs' head pose from the visual modality allows us to identify the relevant passages.

To build our measures, we need to align twice. For the MPs' average pitch per word, we explicitly align the text and the audio modality. Since the audio modality is continuous per default, we identify discrete elements with transcription. Extracting each word's exact start and end time using \textit{WhisperX} \autocite{bain2023whisperx}, we use the timestamps to connect the audio and the text modality. We then retrieve the pitch of each word with the open-source audio analysis software \textit{librosa} \parencite{mcfee2015librosa} and compute deviations from the MP specific pitch averages for each MP.\footnote{Please see Appendix \ref{pitch_est} for more details on measuring the speakers' mean pitch.} 

We explicitly align the continuous visual and audio modalities to understand when MPs address the AfD. In the visual modality, the elements of interest are those moments when MPs look at the far right. Technically, we first need to determine the passages that record MPs from the frontal perspective.\footnote{This is important due to multiple camera perspectives. Please see Appendix \ref{pitch_est} for more details on how we detect the speaker.} We then measure the head pose from left to right with the convolutional neural net implemented in the python package \textit{SixDRepNet} \autocite{Hempel_2022}. In the seating arrangement of the 19th \textit{Bundestag}, the AfD sits between 45\degree\space and 70\degree\space relative to the speaker at the podium.\footnote{Please see Appendix \ref{pitch_est} for more details on measuring the direction of speakers' gaze.} Knowing the angle of the MP's head pose, we can thus annotate the passages when an MP addresses the AfD. Again, we rely on timestamps to connect non-verbal information about attention to the AfD with the corresponding information about the pitch per word from the audio modality.

\begin{figure}
    \centering
    \includegraphics[width=.8\textwidth]{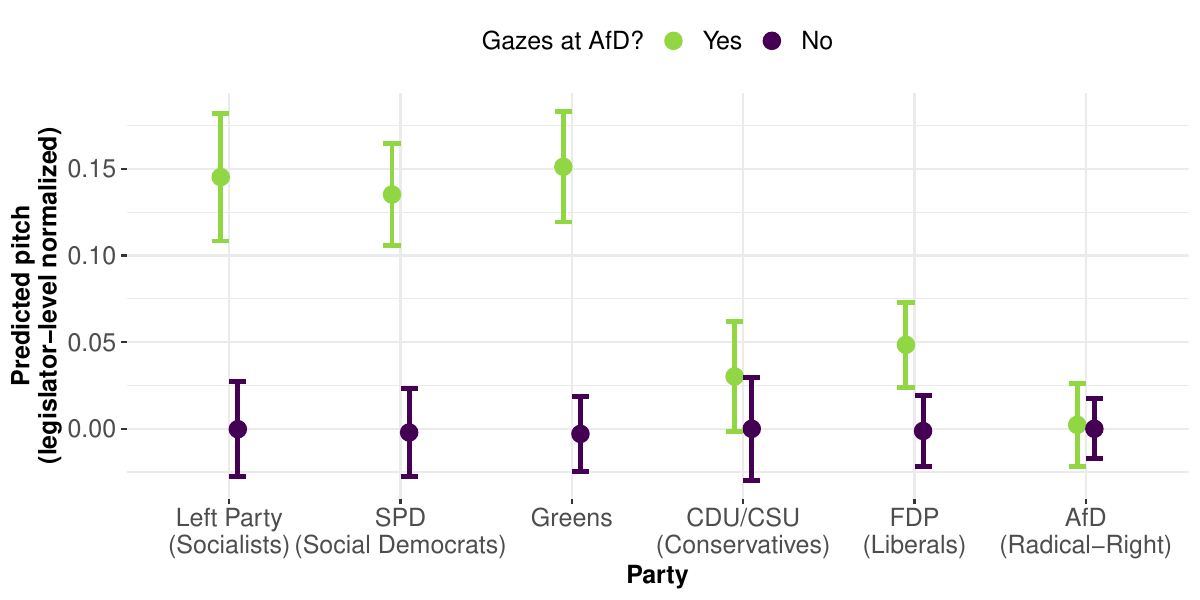}
    \caption{Predicted MPs arousal (standardized audio pitches) with 95\% confidence intervals when addressing the AfD (green), and when addressing all other parties (purple) during the 19th German \textit{Bundestag}.}
    \label{fig:afd-pitch}
\end{figure} 

Once all modalities are aligned, we address the research question with a regression model. The standardized pitch is the dependent variable; the interaction between the party and a dummy variable that indicates whether an MP addresses the AfD are the key explanatory factors. We additionally include fixed effects for MPs for statistical control. Figure~\ref{fig:afd-pitch} charts each faction's predicted audio pitch on the vertical axis. Values in green indicate when MPs address the AfD, and values in purple show when MPs speak to anybody else. For all parties, MPs' predicted pitch is slightly lower than the standardized mean at 0, a value we would expect since these parts of the speeches represent the large majority of the corpus. However, the green values are of core interest for our research question since they capture whether MPs are more upset when talking to the AfD. This is the case, and in particular so for the three left factions from the socialist \textit{Die Linke}, the social democrats \textit{SPD}, and the green party \textit{Bündnis 90/Grüne} (0.15 standard deviation higher pitch). The MPs from the conservative faction from \textit{CDU} and \textit{CSU} show a higher pitch, too (0.03 standard deviation), just as MPs from the liberal party \textit{FDP} (0.05 standard deviation). There is no difference in the pitch level when MPs from the AfD talk to their faction. Our findings align with theoretical expectations regarding communication in the 19th \textit{Bundestag}.

The main goal of this application is to show how systematic, explicit alignment of modalities can address the connectedness of multimodal data. Alignment prepares the multimodal data so that its information can be analyzed with a dedicated regression model. Mapping the modalities also highlights the potential of cross-modal queries, which add significant analytical leverage to research designs in multimodal analysis. It would be possible to look for any other distinct semantic communication symbol that can be easily detected, such as facial expressions or other body language in the visual modality, sounds (e.g., bells or applause) in the audio modality \autocite{AshForthcoming}, or particular meanings (e.g., populist sentences, reference to historical events) in the text modality.

\section{Conclusion}\label{conclusion}

Multimodal data offers a powerful lens for political analysis, reflecting the complexity of human communication more accurately than any single modality alone. However, this richness comes with significant analytical challenges. Differences in data representation, the conceptual levels at which information aligns, and the ways in which signals interact all add layers of complexity. To fully harness the potential of multimodal data, researchers must address these challenges systematically. Alignment is a key step in this process, ensuring that information from one modality is meaningfully related to another before its analysis.

We propose three guiding questions to help researchers determine the appropriate alignment strategy: (1) Is the data continuous, or does it contain discrete semantic elements? (2) If discrete, does the data represent these elements in a semantic or non-semantic way? (3) Should alignment be handled as a separate pre-processing step or as integrated into the model itself? Our applications\textemdash predicting the tonality of U.S. presidential campaign ads and enabling cross-modal queries in German parliamentary debates\textemdash demonstrate the benefits of alignment in improving model performance and expanding analytical capabilities.

Despite its promise, our literature review reveals that multimodal data remains underutilized in political science. Given the widespread recognition of its potential, this gap is surprising, yet it underscores the challenges of multimodal modeling. Implementing robust methods for integrating diverse data types remains a formidable task, and alignment is a crucial component in overcoming these obstacles. We hope our decision heuristic will serve as a practical guide, lowering the barriers to multimodal research and encouraging more scholars to incorporate these rich data sources into their work.

\newpage
\newpage
\printbibliography
\appendix
\setcounter{table}{0}
\renewcommand{\thetable}{A\arabic{table}}
\setcounter{figure}{0}
\renewcommand{\thefigure}{A\arabic{figure}}

\newpage
\begin{refsection}
\setcounter{page}{1}

\begin{center}
    \begin{LARGE}
    \textbf{Alignment Helps Make the Most of Multimodal Data\\}
    \end{LARGE}
    Christian Arnold (University of Birmingham)\\ Andreas Küpfer (Technical University of Darmstadt)
    \vspace{1cm}

 Appendix
\end{center}

\section*{Table of Contents}

\begin{itemize}
    \item[A.] Literature Review
    \item[B.] Application: Alignment to Improve the Prediction of Tone in Campaign Videos
    \begin{itemize}
        \item[B.1.] Hyperparameter Tuning Grid
        \item[B.2.] Results of Implicit Alignment
    \end{itemize}
    \item[C.] Application: Alignment for Cross-Modal Queries in Parliamentary Speeches
    \begin{itemize}
        \item[C.1.] Preprocessing Strategy
        \item[C.2.] Regression Results
        \item[C.3.] Fightin’ Words
    \end{itemize}
\end{itemize}

\newpage
\renewcommand{\thesection}{\Alph{section}}

\section{Literature Review on Multimodal Data Alignment Papers}

We systematically classified all relevant papers published in six major political science journals (AJPS, APSR, BJPS, JOP, PA, and PolCom) between 2019 and 2024 to show the current state of multimodal data alignment in the discipline. On 25 October 2024, we crawled the database of Google Scholar with the Python package \texttt{scholarly} \autocite{cholewiak2021scholarly}. We applied the search string \texttt{multimodal OR video OR audio OR text OR image} and a filter restriction for year and journal. Our search string is relatively broad to avoid overlooking any relevant paper, so our initial database consists of 2,703 publications. We then manually annotated this corpus to reduce the set of potential candidates. We split our annotation process into two steps. In step one, we only keep papers that study data quantitatively and, more importantly, are suspected to incorporate more than one modality. At the end of this initial manual annotation, we identified 115 papers.

In the second step, we classified the remaining papers into the following three categories that indicate if and how the authors of the papers apply multimodal alignment. We only considered each paper's main body and Appendix for our annotation. There are two decisions to make: First, if the paper uses multimodal data. Second, if the authors implement multimodal alignment. For the former question, we ask the following:

\begin{itemize}
    \item[1] Do the authors model multimodal data?
    \begin{itemize}
        \item[1.1] If so: What are the different modalities?
    \end{itemize}
\end{itemize}

Once we know that a paper is working with multimodal data, we seek to understand whether the authors align the modalities. Again, we annotate the answers to a range of questions:

\begin{itemize}
    \item[2] Do the authors define elements in the modalities?
        \begin{itemize}
            \item[2.1] If so: What are the elements in each modality?
        \end{itemize}
    \item[3] Do the authors define relationships or correspondences between elements?
        \begin{itemize}
            \item[3.1] If so: How do they define the relationships or correspondences?
            \item[3.2] If so: How do they align the relationships or correspondences? I.e., do they align via model or the data?
        \end{itemize}
\end{itemize}

\begin{sidewaystable}[htb]
\begin{center}
\caption{Overview of the subset of papers annotated as studying multimodal data in AJPS, APSR, BJPS, JOP, PA, and PolCom between 2019 and 2024.}
\begin{tabular}{l l l p{4.5cm} p{4cm}}
\hline
Article & Journal & 1 (1.1) & 2 (2.1) & 3 (3.1/3.2)\\
\hline
Dietrich \autocite*{Dietrich_2021}              & PA    &  \checkmark\hspace{.25em} (Video/Video) & \checkmark\hspace{.25em} (Video frames that (do not) show overhead shot) & \checkmark\hspace{.25em}(Overhead shot clip/Explicit)   \\
Dietrich, Enos, and Sen \autocite*{dietrich2019emotional}            & PA    & \checkmark\hspace{.25em} (Text/Audio) & \checkmark\hspace{.25em} (Text dictionary/Audio utterance-segments) & \xmark\hspace{.25em}     \\
Kaufman and Klevs \autocite*{kaufman2022adaptive}              & PA & \checkmark\hspace{.25em} (Text/Text) & \checkmark\hspace{.25em} (Letters) & \xmark\hspace{.25em} \\
Licht \autocite*{Licht_2023}                       & PA  & \checkmark\hspace{.25em} (Text/Text) & \checkmark\hspace{.25em} (Sentence embeddings) & \checkmark\hspace{.25em}(Self-attention/Implicit)       \\
Mozer et al. \autocite*{mozer2020matching}                & PA  & \checkmark\hspace{.25em} (Text/Text) & \checkmark\hspace{.25em} (Words) & \xmark\hspace{.25em}       \\
Tarr, Hwang, and Imai \autocite*{tarr2022automated}                & PA  & \checkmark\hspace{.25em} (Text/Video) & \checkmark\hspace{.25em} (Text and video OCR mentions of issue keywords) & \xmark\hspace{.25em} \\
Boussalis et al. \autocite*{boussalis2021gender}              & APSR & \checkmark\hspace{.25em} (Text/Audio/Video) & \checkmark\hspace{.25em} (Text on statement-level/Audio in one-second-segments/Video each face in each frame) & \xmark\hspace{.25em}       \\
Dietrich, Hayes, and O’Brien \autocite*{dietrich2019pitch}                & APSR   & \checkmark\hspace{.25em} (Text/Audio) & \checkmark\hspace{.25em} (Words related to 'female issues'/Audio high levels of pitch) & \xmark\hspace{.25em}    \\
Fowler et al. \autocite*{fowler2021political}              & APSR   & \checkmark\hspace{.25em} (Text/Audio/Video) & \checkmark\hspace{.25em} (Text and audio word-level/Video one frame every 15 seconds) & \xmark\hspace{.25em}     \\
Rittmann \autocite*{rittmannforthcominglegislators}   & BJPS   & \checkmark\hspace{.25em} (Text/Audio) & \checkmark\hspace{.25em} (Words related to 'female issues'/Audio high levels of pitch)  & \xmark\hspace{.25em}     \\
Boussalis and Coan \autocite*{boussalis2021facing}              & PolCom   & \checkmark\hspace{.25em} (Text/Video) & \checkmark\hspace{.25em} (Text words aggregated to sentences/Video frames happy and unhappy expression) & \xmark\hspace{.25em}      \\
Fox \autocite*{fox2023ethnic}                    & PolCom   & \checkmark\hspace{.25em} (Text/Image) & \checkmark\hspace{.25em} (Implicit annotation of objects that stand for ethnicities) & \xmark\hspace{.25em}     \\
Sang Jung Kim and Chen \autocite*{kim2023vaccinetiktok}             & PolCom     & \checkmark\hspace{.25em} (Text/Video) & \checkmark\hspace{.25em} (Video faces/Text dictionary) & \xmark\hspace{.25em}    \\

\hline
\end{tabular}
\label{table:lit_review}
\end{center}
\end{sidewaystable}

Table \ref{table:lit_review} gives an overview of our annotations. We find 13 papers that model multimodal data. Most of these papers combine discrete (text) and continuous (video or audio) modalities except for three articles that model two textual data streams \autocite{kaufman2022adaptive, Licht_2023, mozer2020matching} and one that studies how text relates to images \autocite{fox2023ethnic}. While all authors define elements in modalities, only two connect these elements. Dietrich \autocite*{Dietrich_2021} defines them by explicitly aligning for sequences in plenary debates that show overhead shots of Members of Congress in the US. Licht \autocite*{Licht_2023} uses implicit alignment for multilingual classification of political text. He aligns the sentence embeddings of different languages, leveraging the self-attention mechanism of transformer models. Finally, this subset of research only shows papers that decided to model multimodal data, and we do not know how many authors decided against the use of multimodality in their projects due to its complexity of modelling.

\section{Application: Aligning Heterogeneous Data to Improve the Prediction of Tone in Campaign Videos}

\subsection{Preprocessing Strategy}\label{preprocess_strat}

As we generate the text modality with \textit{WhisperX}, we also transform the audio and video modality with modality-tailored techniques. Applying \textit{pyAudioAnalysis} \autocite{giannakopoulos2015pyaudioanalysis}, we extract 34 typical audio features for every 0.1 seconds with an overlap of 0.05 seconds from the audio track. Similarly, we rely on \textit{Py-Feat}\footnote{https://py-feat.org/} to identify 170 facial features on the video modality. If more than one face is visible on a single frame, we calculate the mean of the features across all faces for that frame. We only keep every 8th audio window and video frame to keep the resulting data manageable for our models.

\subsection{Hyperparameter Tuning Grid}

\begin{table}[h]
\centering
\caption{Hyperparameter ranges for unimodal and multimodal model training.}
\label{tab:hyperparameter_ranges}
\begin{tabular}{ll}
\toprule
\textbf{Hyperparameter} & \textbf{Values} \\
\midrule
Number of Epochs & \{20, 30\} \\
Number of Attention Heads & \{5, 10\} \\
Number of Kernels (Language) & \{1, 3, 5\} \\
Number of Kernels (Vision) & \{1, 3, 5\} \\
Number of Kernels (Audio) & \{1, 3, 5\} \\
Dropout (Attention) & \{0.1, 0.3\} \\
Clip & \{0.8, 1\} \\
\bottomrule
\end{tabular}
\end{table}

\subsection{Results of Implicit Alignment}

\begin{table}[h]
\centering
\caption{Detailed F1 scores of tone classification per seed in unimodal and multimodal scenarios.}
\label{tab:appendix_results}
\begin{tabular}{cp{3cm}p{3cm}lll}
\toprule
\textbf{Seed} & \multicolumn{2}{c}{\textbf{Multimodal F1}} & \multicolumn{3}{c}{\textbf{Unimodal F1}}\\
\cmidrule(lr){2-3} \cmidrule(lr){4-6}
& \textbf{Cross-modal Attention} & \textbf{No cross-modal Attention} & \textbf{Text} & \textbf{Audio} & \textbf{Video} \\
\midrule
20230701 & 0.955 & 0.940 & 0.932 & 0.849 & 0.853 \\
20230702 & 0.934 & 0.908 & 0.934 & 0.849 & 0.855 \\
20230703 & 0.934 & 0.905 & 0.923 & 0.847 & 0.857 \\
20230704 & 0.965 & 0.936 & 0.946 & 0.862 & 0.877 \\
20230705 & 0.935 & 0.911 & 0.917 & 0.850 & 0.836 \\
\bottomrule
\end{tabular}
\end{table}

\section{Application: Alignment for Cross-Modal Queries in Parliamentary Speeches}

\subsection{Preprocessing Strategy}\label{pitch_est}

To identify whether the speaker is visible in a speech video, we train the Detection Transformer (\textit{DETR}) framework \autocite{carion2020endtoend} to recognize the special lettering \textit{Deutscher Bundestag} as well as Germany's coat of arms (\textit{Bundesadler}) on the speaker's desk.\footnote{We use transfer learning with a pre-trained \textit{ResNet-50} and fine-tuned with a few annotated video frames. Our final model successfully recognizes the combination of lettering and logo in 95.00\% of the images in our validation set.} 

Estimating the standardized vocal pitch of all speakers is straightforward. In light of differences in frequency ranges for men and women, we follow the range settings proposed by Dietrich, Hayes, and O’Brien \autocite*{dietrich2019pitch} and Rittmann \autocite*{rittmannforthcominglegislators} and set the floor pitch to 75Hz and its ceiling to 300Hz for men and to 100Hz to 500Hz for woman. Lastly, we normalize all values per MP as standardized word-wise differences from their mean.

\textit{SixDRepNet} allows us to extract the gaze direction of a speaker.\footnote{The mean absolute error of the angle measurement is 3.72 \autocite{Hempel_2022}.} In addition of our \textit{yaw} setting between 45\degree and 70\degree, we consider \textit{pitch} angles smaller than -20\degree to correct when an MP looks at their speech notes and keeps the measurement to the AfD active during that time. We also require a minimum length of 10 words to identify a part of the speech to be directly addressed to the AfD.

\subsection{Regression Results}

\begin{table}[h!]
\begin{center}
\caption{Regression model based on 6'723'156 words in all plenary debates of the 19th legislative period of the German \textit{Bundestag}. Fixed effects on the ID of an MP.}
\begin{tabular}{l c}
\hline
 &  Fixed Effects GLM \\
\hline
Looks at AfD                                  & $0.00$        \\
                                              & $(0.02)$      \\
Party: CDU/CSU (Conservatives) x Looks at AfD & $0.03$        \\
                                              & $(0.04)$      \\
Party: Left (Socialists) x Looks at AfD       & $0.14$        \\
                                              & $(0.05)$      \\
Party: FDP (Liberals) x Looks at AfD          & $0.05$        \\
                                              & $(0.04)$      \\
Party: Greens x Looks at AfD                  & $0.15$        \\
                                              & $(0.05)$      \\
Party: SPD (Social Democrats) x Looks at AfD  & $0.14$        \\
                                              & $(0.04)$      \\
\hline
Num. obs.                                     & $6723156$     \\
Num. groups: MP ID                            & $695$         \\
Deviance                                      & $6721773.29$  \\
Log Likelihood                                & $-9539054.69$ \\
Pseudo R$^2$                                  & $-0.00$       \\
\hline
\end{tabular}
\label{table:coefficients}
\end{center}
\end{table}

\subsection{Fightin' Words}

Another research question of our application identifies the words MPs choose when they address the AfD.
Querying text based on information is again a matter of using information on the video domain based on time stamps. This time, however, the query retrieves information from the text modality and not from the audio modality.

To better understand the communication patterns between the faction from the AfD and all others, we use the Fightin' Words algorithm \autocite{monroe_colaresi_quinn_2008}. It compares word frequencies across multiple groups, considering the frequency of each word in the entire dataset. In each of the four possible speaking situations (others $\rightarrow$ AfD, AfD $\rightarrow$ AfD, others $\rightarrow$ others, AfD $\rightarrow$ others), the y-axis indicates how strongly associated they are with talking to the respective audience. The x-axis depicts how frequently the words are used in general. 

\begin{figure}[t]
    \centering
    \includegraphics[width = .95\textwidth]{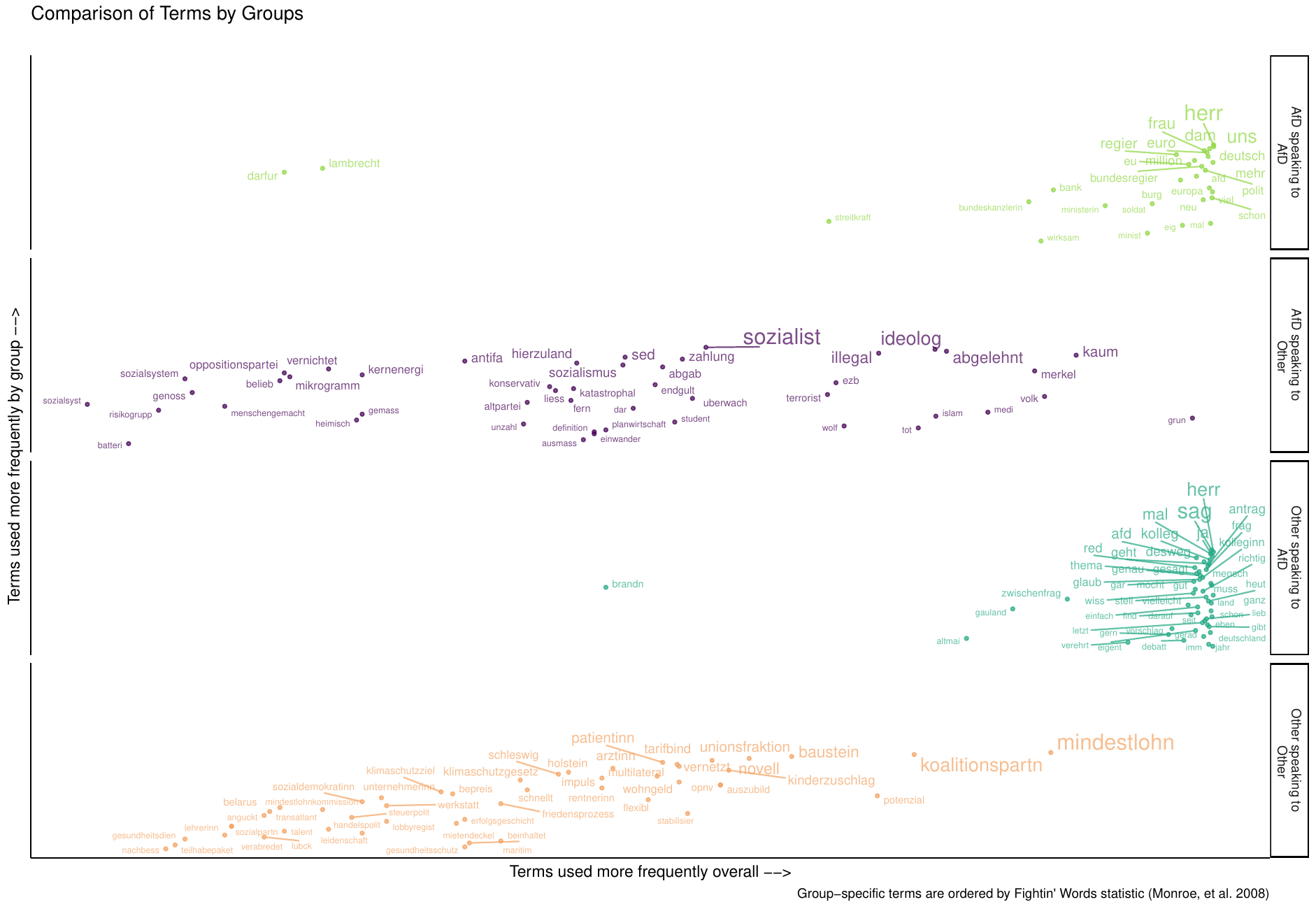}
    \caption{Fightin' Words algorithm \autocite{monroe_colaresi_quinn_2008} to analyze the frequency of words in four possible scenarios of an MP (others $\rightarrow$ AfD, AfD $\rightarrow$ AfD, others $\rightarrow$ others, AfD $\rightarrow$ others).}
    \label{fig:afd_words}
\end{figure}

Figure \ref{fig:afd_words} presents our results for all four scenarios. Whenever other parties talk to the AfD, they typically do not use their speech to engage with the faction thematically. Instead, they rely on words that are more indicative of a rebuttal of an interjection, such as \textit{verehrt} (honoured), \textit{herr} (Mr.), \textit{gauland} (Alexander Gauland (AfD)), \textit{kolleg} (word stem of colleague), \textit{Antrag} (proposal), \textit{Frage} (question). In short, the speaker's audience is correlated to the spoken words, and MPs do not seem to engage with the AfD on the grounds of particular contents but rather to reprimand the faction.

\newpage

\printbibliography
\end{refsection}

\end{document}